\documentclass[letterpaper, 10 pt, conference]{ieeeconf}  

\IEEEoverridecommandlockouts                              

\usepackage{amsmath,amssymb,amsfonts}
\usepackage[linesnumbered,algoruled,boxed,lined]{algorithm2e}
\SetKwInput{kwInit}{Initialize}
\let\labelindent\relax \usepackage[inline]{enumitem}
\usepackage{graphicx}
\usepackage[table,xcdraw]{xcolor}
\usepackage{soul}
\usepackage{xfrac}
\usepackage{cite}
\usepackage{booktabs}
\usepackage[caption=false]{subfig}
\usepackage[colorlinks=true,linkcolor=blue]{hyperref}
\DeclareRobustCommand{\bigO}{%
  \text{\usefont{OMS}{cmsy}{m}{n}O}%
}
\newtheorem{remark}{Remark}
\newcommand{\T}{\top}

\title{\LARGE \bf
Frenet Corridor Planner: An Optimal Local Path Planning Framework for Autonomous Driving
}

\author{Faizan M. Tariq$^{1}$, Zheng-Hang Yeh$^{1}$, Avinash Singh$^{1}$, David Isele$^{1}$, Sangjae Bae$^{1}$
\thanks{$^{1}$Honda Research Institute, USA. Email: {\tt\small \{faizan\_tariq, zheng-hang\_yeh, avinash\_singh, disele, sbae\}@honda-ri.com}}
}

\begin{document}

\maketitle
\thispagestyle{empty}
\pagestyle{empty}

\begin{abstract}
Motivated by the requirements for effectiveness and efficiency, path-speed decomposition-based trajectory planning methods have widely been adopted for autonomous driving applications. While a global route can be pre-computed offline, real-time generation of adaptive local paths remains crucial. Therefore, we present the Frenet Corridor Planner (FCP), an optimization-based local path planning strategy for autonomous driving that ensures smooth and safe navigation around obstacles. Modeling the vehicles as safety-augmented bounding boxes and pedestrians as convex hulls in the Frenet space, our approach defines a drivable corridor by determining the appropriate deviation side for static obstacles. Thereafter, a modified space-domain bicycle kinematics model enables path optimization for smoothness, boundary clearance, and dynamic obstacle risk minimization. The optimized path is then passed to a speed planner to generate the final trajectory. We validate FCP through extensive simulations and real-world hardware experiments, demonstrating its efficiency and effectiveness.
\end{abstract}


\section{Introduction}
Path-speed decomposition approaches are widely used in autonomous vehicle trajectory planning (path and speed) due to their reliability and efficiency \cite{kant1986toward,fraichard1993dynamic,anon2024mpqp}. These methods simplify the overall trajectory planning problem by separately handling path and speed planning, resulting in computationally efficient algorithms \cite{koenig2002d,likhachev2005anytime,multifuture}. Building on this framework, this work focuses on developing an efficient path planning strategy for autonomous driving.

\begin{figure}
    \centering
    \includegraphics[width=0.8\columnwidth]{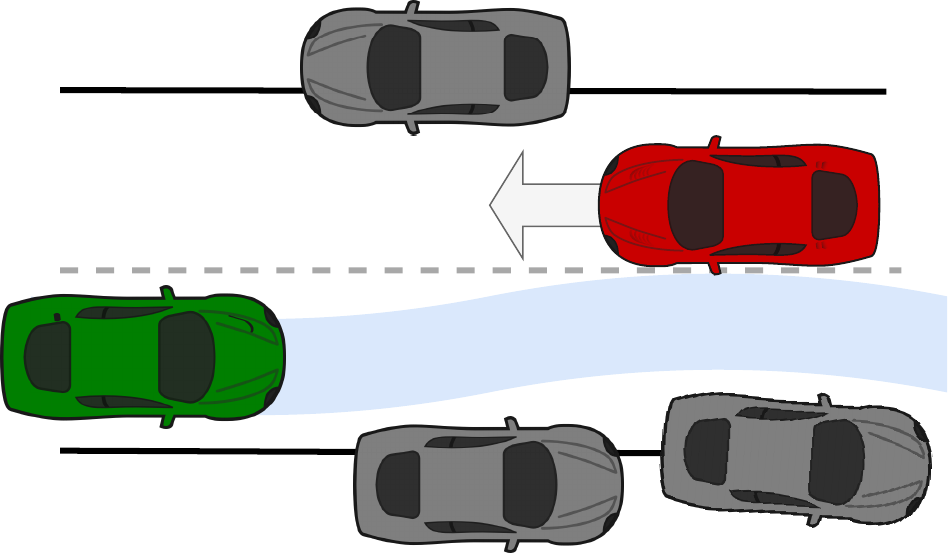}
    \caption{\textbf{Motivational scenario}. The ego vehicle (in green) must deviate from the lane center to avoid a collision with the parked cars on the roadside while being cognizant of the oncoming traffic. Without a local path planner, the ego vehicle may remain stuck, waiting indefinitely for the parked cars to move before proceeding along its pre-determined global route.}
    \label{fig:motivation}
    \vspace{-15pt}
\end{figure}

Consider the scenario illustrated in Fig.~\ref{fig:motivation}. If a route were generated in advance from the map data, it would likely not incorporate the cars parked on the side of the road. However, deviating around the parked cars must be done in a way that minimizes disruption to the oncoming traffic. This necessitates online path generation where the path can be updated smoothly while considering the dynamic limitations of the ego vehicle.

\subsection*{Related Work}
There is an extensive history of path planning research including learning-based approaches \cite{huang2023neural, anjian_consistencyModel}, artificial potential field methods \cite{hwang1992potential}, graph search techniques \cite{A*,daniel2010theta}, sampling-based methods \cite{RRT*,jordan2013optimal}, polynomial (parametric) optimization strategies \cite{xu2012real}, and non-parametric optimization methods \cite{dirckx2023optimal}. 
However, much of this work does not simultaneously address kinematic feasibility, path smoothness, and computational efficiency for time-sensitive and safety-critical applications, such as autonomous driving.

Formal optimization methods can directly incorporate dynamic and safety constraints, but this often comes at a detriment to computational efficiency. Non-linear vehicle dynamics \cite{kong2015kinematic, rcms}, the existence of multiple routes through space-time \cite{phan2020covernet,salzmann2020trajectron++}, and the uncertainty and interactivity of other traffic participants \cite{sadigh2016planning,ma2021reinforcement, bae2022lane,hu2023active, bidirectional_overtaking} all combine to greatly increase the complexity of the problem. This results in a highly non-convex formulation which takes significant computation time and resources to solve. Our approach follows this line of work but focuses on systematically reducing complexity by dividing the problem into stages, achieving efficient path planning at a low computational cost.

\subsection*{Contribution}
In this work, we propose the Frenet Corridor Planner (FCP), an efficient optimization-based path planning strategy that generates smooth paths around obstacles through a multi-stage process. First, vehicles in the environment are represented as safety-augmented bounding boxes, while pedestrian clusters are modeled as convex hulls in the Frenet space. The appropriate deviation side is then determined for each static obstacle, and the drivable region (corridors) for the ego vehicle is established. Next, an optimization problem is solved to generate a path that maximizes smoothness, maintains a safe distance from corridor boundaries, and aligns with the reference path while considering the risk associated with dynamic obstacles. This is achieved using a modified space-domain bicycle kinematics model, which, to the best of our knowledge, has not been previously explored. Finally, the generated path is passed to the speed planner to produce the overall trajectory. Our approach is rigorously evaluated in the presence of perception noise through both simulations and physical hardware experiments.

\section{Problem Setting}
\subsection{Frenet Coordinate System}   \label{sec:frenet}
In this work, we utilize the Frenet Coordinate System \cite{frenet1852courbes}, where the $s$-axis is representative of the longitudinal displacement along a given reference (global) path ($\mathbf{ref}$), while the $d$-axis denotes the lateral displacement orthogonal to $\mathbf{ref}$. Any point $\alpha \in \mathbb{R}^2$ can be transformed from the Cartesian to the Frenet frame using a non-linear transformation as follows: 
\begin{equation}    \label{eq:frenet-transform}
    f_c \colon (\alpha_x, \alpha_y, \mathbf{ref}) \mapsto (\alpha_s, \alpha_d).
\end{equation}

\subsection{System Architecture}    \label{sec:system-architecture}
The overall modular system architecture is adopted from our previous work \cite{tariq2022slas}, with this study focusing specifically on the motion planning layer. We tackle the trajectory planning problem with a decoupled scheme (Fig. \ref{fig:pipeline}) that allows us to tackle the path and speed planning problems independently. This decomposes the overall problem complexity to yield an efficient trajectory planning algorithm. In our previous work, we developed a robust speed planning method \cite{anon2024mpqp}, so this work serves to fill in the gap by proposing a corresponding path planning scheme. 
The generated trajectory, consisting of both path and speed, is forwarded to the downstream Control module, which also utilizes a decoupled control scheme \cite{bae2022lane}. Longitudinal control is managed by a PID controller, tuned using the CARLA simulator \cite{carla}, while lateral control is handled by a model predictive controller (MPC) designed with a one-time-step planning horizon based on the vehicle rotation model \cite{vehicleRotationModel}.



\begin{remark}
Although we have used our previously developed algorithms for the downstream speed and control modules in the validation studies (Section~\ref{sec:validation}), the modular system architecture provides the flexibility to integrate any external algorithm from the literature.
\end{remark}

\subsection{Planning Pipeline}
\begin{figure}
    \centering
    \includegraphics[width=0.95\columnwidth]{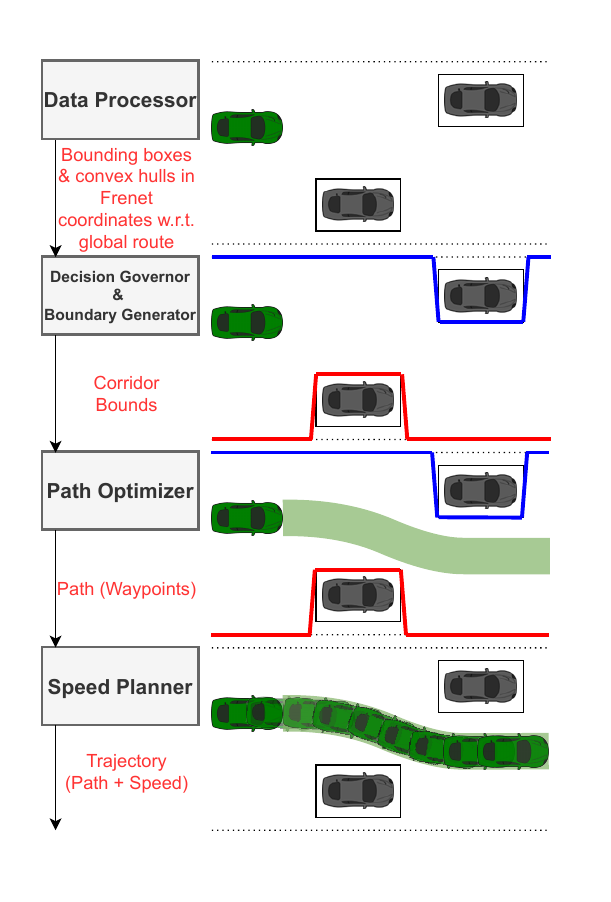}
    \caption{\textbf{Trajectory planning pipeline}. The data flow between the various building blocks of FCP is illustrated on the left, while the output visualization from each module is shown on the right.}
    \label{fig:pipeline}
    \vspace{-15pt}
\end{figure}

The Frenet Corridor Planner (FCP) in itself consists of several submodules that work together to generate a path for the downstream speed planning and control modules to follow, as depicted in Fig. \ref{fig:pipeline}. The Data Processor (DP) processes perception and localization data to extract obstacle-related information in the Frenet frame, which is then sent to the Decision Governor (DG). The DG determines the appropriate side to deviate each obstacle and forwards this information to the Boundary Generator (BG). The BG defines the boundaries of the drivable region (corridor) and transmits them to the Path Optimizer (PO), which generates a path using our proposed space domain kinematics model.

\section{Frenet Corridor Planner}
\subsection{Data Processor (DP)}
The DP takes in the obstacles' state (position, orientation, and size) information from the external perception and localization module \cite{tariq2022slas} to generate safety-augmented bounding boxes for vehicles and convex hulls for pedestrian clusters with the help of DBSCAN \cite{dbscan}, a density-based clustering algorithm. Any pedestrian not associated with a cluster is treated as an independent obstacle. At the current time $t$, the obstacle set containing the linearly interpolated points along the edges of the bounding boxes and convex hulls in the Frenet frame is denoted by $\mathcal{O}_t$.

\subsection{Decision Governor (DG)}
In addition to having information on the locations and speeds of the obstacles present in the environment, FCP requires further information on the appropriate side of deviation for a static obstacle. Specifically, FCP needs to know whether a static obstacle should be considered in the upper or lower boundary of the corridor, and this information is provided by the DG. Therefore, DG partitions the obstacle set ($O_t$) into disjoint lower ($\mathcal{O}_t^\text{lb}$) and upper ($\mathcal{O}_t^\text{ub}$) bound obstacle sets such that $O_t = \mathcal{O}_t^\text{ub} \cup \mathcal{O}_t^\text{ub}$ and $\mathcal{O}_t^\text{ub} \cap \mathcal{O}_t^\text{ub} = \emptyset$.

\begin{figure}
    \centering
    \includegraphics[width=\columnwidth]{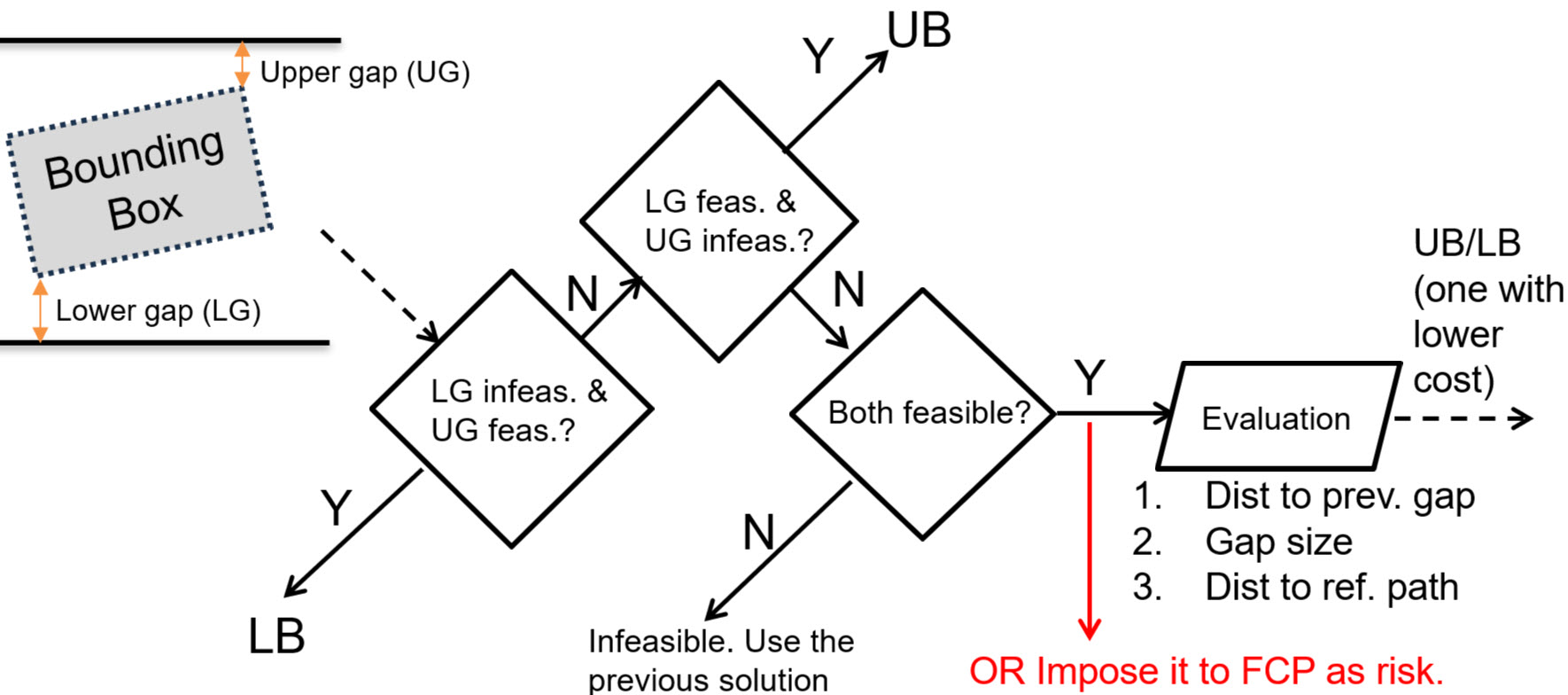}
    \caption{\textbf{Decision tree for boundary classification of each obstacle}. The decision tree evaluates the lower and upper gaps within the drivable space. If both gaps are available, two approaches can be used: selecting the preferred gap based on cost evaluation or treating the obstacle as a risk in PO.}
    \label{fig:decision-governor}
\end{figure}

Since this work focuses primarily on the path planning layer, we demonstrate the efficacy of FCP with a simple decision tree \cite{song2015decision} for our DG, as shown in Fig.~\ref{fig:decision-governor}. 
This simple decision tree performs well even with moderate noise in the perception and localization data. However, with large noise, it may cause inconsistent obstacle classification (i.e., switching between lower and upper bounds), especially when an obstacle is located near the center of the drivable space. Alternatively, the obstacle can be treated as a risk in the PO (see Section~\ref{sec:dynamic-obstacle-handling}) to enhance path consistency.
Owing to the modularity of our approach, more advanced methods from the literature \cite{planningSurvey} can be incorporated into DG to improve the robustness of the trajectory planning pipeline.

\subsection{Boundary Generator (BG)} \label{sec:boundary-generator}
\begin{figure}
    \centering
    \includegraphics[width=\columnwidth]{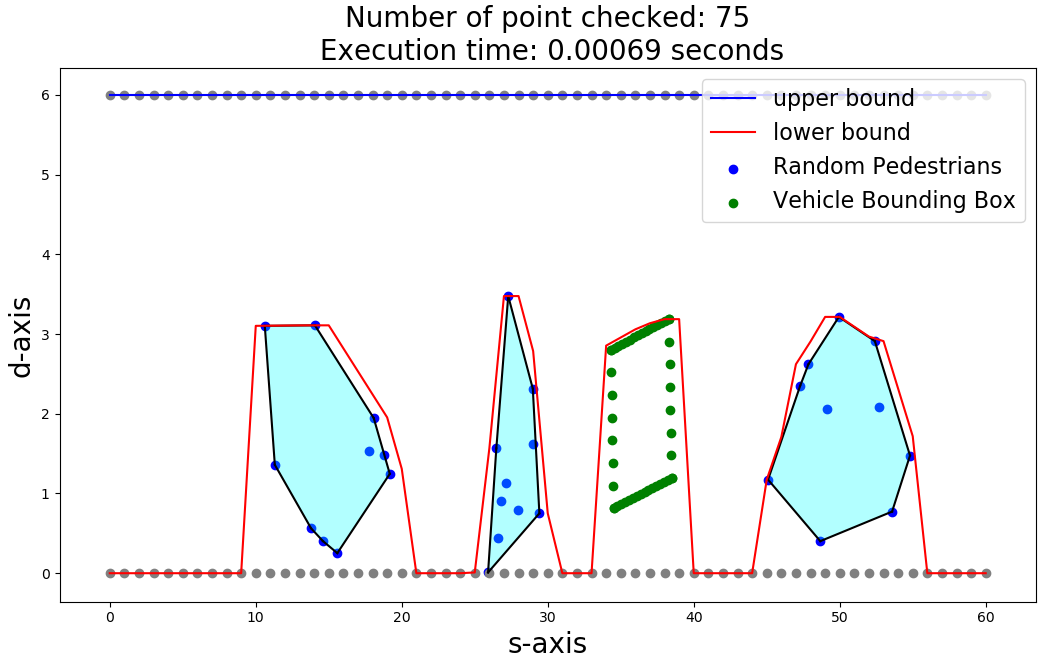}
    \caption{\textbf{Boundary Generation}. With the pedestrians shown as blue dots, the augmented vehicle boundary depicted by green dots, and the convex hulls of pedestrian clusters given by the blue lines, Algorithm~\ref{alg:boundary_generation} generates the lower and upper bounds, shown by the red and blue lines, respectively.}
    \label{fig:test-get-bounds}
    \vspace{-15pt}
\end{figure}

The BG takes in $\mathcal{O}_t^\text{lb}$ and $\mathcal{O}_t^\text{ub}$ as well as the lower and upper road limits, $l_t^\text{lb}$ and $l_t^\text{ub}$, to generate the lower and upper bounds, $d_t^\text{lb}$ and $d_t^\text{ub}$, in the Frenet space for a planning horizon of $N$ steps with a longitudinal spacing of $\Delta s$ meters, resulting in a planning distance of $L=N \times \Delta s$ meters. The boundary generation algorithm, outlined in Algorithm~\ref{alg:boundary_generation}, essentially identifies the ``extremum" d-value for each obstacle point at the queried s-positions. Therefore, the algorithm requires only a single pass over the obstacle points resulting in a runtime complexity of $\bigO (N_o)$ where $N_o$ corresponds to the sum of the number of boundary points for all obstacles in $\mathcal{O}_t$. The output of BG is depicted in Fig. \ref{fig:test-get-bounds}. 






\begin{algorithm}[hbt!] \label{alg:boundary_generation}
\caption{Boundary Generation Algorithm}
\DontPrintSemicolon
\KwIn{$\mathcal{O}_t^\text{lb}$, $\mathcal{O}_t^\text{ub}$, $l_t^\text{lb}$, $l_t^\text{ub}$, $N$, $\Delta s$, $s_0$} 
\KwOut{$d_t^\text{lb}$, $d_t^\text{ub}$}
\kwInit{$d_t^\text{lb} \gets l_t^\text{lb}; d_t^\text{ub} \gets l_t^\text{ub}$}
\vspace{1mm}

\vspace{1mm}

\For{$O \in \{ \mathcal{O}_t\}$}{
    \For{$p = [p_s, p_d] \in O$}{
        $\text{ind} = \left\lfloor \sfrac{(p_s - s_0)}{\Delta s} \right\rfloor$ // Compute $p$ index \;
        \uIf{$O \in \mathcal{O}_t^\text{lb}$}{
        \If{$d_t^\text{lb}[\text{ind}] < p_d$}{
             $d_t^\text{lb}[\text{ind}] = {p_d}$\;
            \If{$\text{ind} \in \{0,\cdots,N-2\}$}{
                $d_t^\text{lb}[\text{ind}+1] = {p_d}$ // Obstacle corners handling \;
            }
        }} \ElseIf{$O \in \mathcal{O}_t^\text{ub}$}{
        \If{$d_t^\text{ub}[\text{ind}] > p_d$}{
             $d_t^\text{ub}[\text{ind}] = {p_d}$\;
            \If{$\text{ind} \in \{0,\cdots,N-2\}$}{
                $d_t^\text{ub}[\text{ind}+1] = {p_d}$ // Obstacle corners handling \;
            }
        }}
    }
}
\end{algorithm}

\subsection{Path Optimizer (PO)}
A path at the current time step $t$ is defined as $P_t = [p_t^1, \cdots, p_t^N]$ where $N$ denotes the number of waypoints and $p_t^i \in \mathbb{R}^2 \ \forall i \in \{1, \cdots, N\}$ denotes a waypoint in space.

\subsubsection{Space Domain Kinematics Model} \label{sec:kinematics}
Since the goal of this work is to generate $P_t$, we transform the vehicle kinematic model from the space-time to space-only domain, allowing us to reduce the problem's complexity. For our base model, we opt for the non-linear kinematic bicycle model, which has been a popular choice for automated vehicle trajectory planning due to its accuracy and efficiency \cite{bicycleModel}.

In the Cartesian Frame, the states, $X_t$, and the control inputs, $U_t$, of the ego vehicle at time instant $t$ are given by
$X_t =
    \begin{bmatrix} 
    x_t & y_t & \psi_t & v_t
    \end{bmatrix}^\T \in \mathbb{X}_t$ and
$U_t = \begin{bmatrix}
    a_t & \delta_t
    \end{bmatrix}^\T \in \mathbb{U}_t$,
respectively. Here, $x_t$, $y_t$, $\psi_t$, and $v_t$ respectively denote the x-coordinate ($m$), y-coordinate ($m$), yaw angle with respect to the x-axis ($rad$), and speed ($m/s$), whereas $a(k)$ and $\delta(k)$ respectively denote acceleration ($m/s^2$), and steering angle ($rad$). The sets $\mathbb{X}_t = \mathbb{R}^2 \times \mathbb{R}_{[0, 2\pi)} \times \mathbb{R}_{[0,V^\text{max}]}$ and $\mathbb{U}_t = \mathbb{R}^2_{[U^{\text{min}}, U^{\text{max}}]}$, respectively, denote the feasible states and the actuation limits. Then, the system dynamics read:
\begin{align}
    x_{t+1} &= x_t + v_t\cos(\psi_t+\beta_t)\Delta t,\\
    y_{t+1} &= y_t + v_t\sin(\psi_t+\beta_t)\Delta t,\\
    \psi_{t+1} &= \psi_t + \frac{v_t}{\ell_r}\sin\beta_t \Delta t,\\
    v_{t+1} &= v_t + a_t \Delta t,
\end{align}
where $t$ denotes the temporal index, $\Delta t$ denotes the sampling timestep, and $\beta_t = \tan^{-1}\left(\frac{\ell_r}{\ell_f+\ell_r}\tan \delta_t\right)$ maps the steering input ($\delta_t$) to the vehicle orientation ($\psi_t$). Now, to remove the time dependence, we fix the distance step as $l_t = v_t\Delta t$.
Moreover, since we explore only the spatial domain, we can disregard the time-varying speed $v_t$ and fix the constant distance step as $l_t \equiv \Delta l$, giving the following modified kinematics model:
\begin{align}
    x_{k+1} &= x_k + \cos(\psi_k+\beta_k)\Delta l,\\
    y_{k+1} &= y_k + \sin(\psi_k+\beta_k)\Delta l,\\
    \psi_{k+1} &= \psi_k + \frac{1}{\ell_r}\sin\beta_k \Delta l,
\end{align}
where $k$ denotes the spatial index. Now, the kinematics model is given in terms of the fixed ``arc length" step along the path, denoted by $\Delta l$.
However, in order to have a uniform correspondence between the path and the upper/lower bounds, we need to further reformulate the kinematics model such that the independent variable is the constant ``longitudinal distance" step ($\Delta s$), instead of $\Delta l$ -- this facilitates the direct integration of $d_t^\text{lb}$ and $d_t^\text{ub}$, generated by the BG, into the PO.

\remark
For the optimization problem (Section \ref{sec:optimization_problem}), the bounds $d_t^\text{lb}$ and $d_t^\text{ub}$ are defined at each ``knot" of the path $P_t$, which necessitates the reformulation of the kinematics model in terms of the independent $\Delta s$ variable.

\remark
With the updated set of states given by $\tilde{X}_k =
    \begin{bmatrix} 
    x_k & y_k & \psi_k
    \end{bmatrix}^\T$,
the modified kinematics model is now governed only by the control input $\delta_k \in \mathbb{R}_{[\delta^\text{min}, \delta^\text{max}]}$.

Now, assuming that we are working in the Frenet frame (Section \ref{sec:frenet}), we can obtain the kinematics in terms of the ``longitudinal distance" step by introducing the following non-linear transformation (projection): $\Delta s = \Delta l \cos(\psi_k+\beta_k)$. The kinematics model then reads:
\begin{align}
    x_{k+1} &= x_k + \Delta s,\\
    y_{k+1} &= y_k + \frac{\sin(\psi_k+\beta_k)}{\cos(\psi_k+\beta_k)}\Delta s\nonumber\\&=y_k + \tan(\psi_k+\beta_k)\Delta s,\\
    \psi_{k+1} &= \psi_k + \frac{\Delta s}{\ell_r}\frac{\sin\beta_k}{\cos(\psi_k+\beta_k)}. \label{eq:heading}
\end{align}
Note that the kinematics model is ill-posed for $\psi_k + \beta_k = \pi/2$. This corresponds to a path $P_t$ orthogonal to $\mathbf{ref}$ -- In the Frenet frame, each point along such a path has the same $s$ value, while the path is not defined for any other $s$ value, leading to a singularity.

\remark The bicycle kinematic model does not translate directly from the Cartesian to the Frenet frame due to a non-linear transformation with respect to $\mathbf{ref}$ (\ref{eq:frenet-transform}). To address this, we introduce a curvature-based model correction in Section \ref{sec:curvature_limits} to ensure conformance of the Cartesian-based kinematic model to the Frenet space.

Now, $\beta_k$ is defined as:
\begin{equation}
    \beta_k = \arctan\left(\frac{l_r}{l_f+l_r}\tan\delta_k\right),
\end{equation}
which is non-linear. To linearize, we approximate:
\begin{equation}
    \beta_k \approx \frac{l_r}{l_f+l_r}\delta_k.
\end{equation}
This approximation is kinematically valid (feasible) only if $\frac{l_r}{l_f+l_r}|\delta_k|$ under-approximates $|\beta_k|$ -- otherwise, the approximated kinematics with the maximum steering angle $\delta_{\max}$ may not be feasible. Thus, we show that $|\beta_k| \geq \frac{l_r}{l_f+l_r}|\delta_k|~\forall \delta_k \in (-\frac{\pi}{2},\frac{\pi}{2})$. For brevity, subscript $k$ is omitted in the following proof. 

\begin{figure}
    \centering
    \includegraphics[width=0.85\columnwidth]{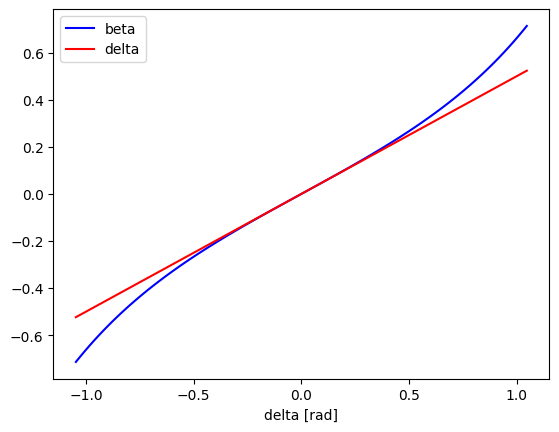}
    \caption{\textbf{Numerical validation for $\mathbf{\frac{l_r}{l_f+l_r}|\delta_k|}$ under-approximating $\mathbf{|\beta_k|}$}. The linear plot $\frac{l_r}{l_f+l_r}\delta_k$ stays below the $\beta_k$ curve for $d_k \in [0,\frac{\pi}{2})$ and above $\beta_k$ for $d_k \in (-\frac{\pi}{2},0]$ showing $|\beta_k| \geq \frac{l_r}{l_f+l_r}|\delta_k|~\forall \delta_k \in (-\frac{\pi}{2},\frac{\pi}{2})$.}
    \label{fig:beta_approx}
    \vspace{-15pt}
\end{figure}

\begin{proof}
Let:
\begin{equation}
    f(\delta) = \arctan\left(a\tan\delta\right) - a\delta,
\end{equation}
where $a = \frac{l_r}{l_f+l_r} \in [0,1]$. We want to show that $f(\delta) \geq 0 \;\forall \delta \in \left[0,\frac{\pi}{2}\right)$ and $f(\delta) \leq 0\;\forall \delta \in \left(-\frac{\pi}{2},0\right]$. The derivative reads:
\begin{align}
    f'(\delta) &= \frac{a\sec^2\delta}{1+a^2\tan^2\delta}-a\\
               &= \frac{a}{\cos^2\delta+a^2\sin^2\delta}-a\\
               &= a\left(\frac{1-(\cos^2\delta+a^2\sin^2\delta)}{\cos^2\delta+a^2\sin^2\delta}\right).
\end{align}
Given $a\in[0,1]$, the following suffices for all $\delta$ (proof is trivial):
\begin{equation}
    \cos^2\delta+a^2\sin^2\delta \leq 1.
\end{equation}
Thus, $f'(\delta) \geq 0$, indicating $f$ is a monotonically increasing function. When $\delta=0$, $f(\delta)=0$. Due to the monotonicity, for $\delta > 0$, $f(\delta) \geq 0$, i.e.,:
\begin{align}
    f(\delta) &= \arctan\left(a\tan\delta\right) - \delta \geq 0\\
    &\longleftrightarrow \arctan\left(a\tan\delta\right) \geq \delta.
\end{align}
\end{proof}
Similarly, one can prove for $\delta\in\left(-\frac{\pi}{2},0\right]$. The numerical validation is shown in Fig.~\ref{fig:beta_approx}. In practice, especially for normal highway driving where the car is not pushed to the actuation limits, the steering range is typically limited to $[-0.6,0.6]$ $rad$ within which the approximation holds.

\subsubsection{Actuation limit induced by reference path's curvature} \label{sec:curvature_limits}
We have thus far transformed the bicycle kinematic model into a space-only kinematic model with longitudinal space sampling. However, we are yet to account for $\mathbf{ref}$'s curvature in the formulation, which is essential for converting the kinematic model from the Cartesian to the Frenet space. Neglecting this curvature can lead to kinematic infeasibility, making it difficult for the vehicle to follow the computed path. To address this, we directly impose curvature limitation on the path-planning problem in Section \ref{sec:optimization_problem} by restricting the feasible actuation set using the angle changes along $\mathbf{ref}$ at each step $k$, given as $\Delta \bar\psi_k$.

Based on the kinematics model derived in Section~\ref{sec:kinematics}, the change in heading angle (\ref{eq:heading}) at any step $k$ reads:
\begin{equation}
    \Delta \psi_k = \frac{\Delta s}{\ell_r}\frac{\sin{u_k}}{\cos{(\psi_k + u_k)}},
\end{equation}
which is lower bounded by $\frac{\Delta s}{\ell_r}\tan{u_k}$ for $|\psi_k + u_k| < \frac{pi}{2}$ where $u_k = \frac{l_r}{l_f+l_r}\delta_k$. We approximate $\Delta \bar\psi_k \approx \frac{\Delta s}{\ell_r}\tan{u_k}$ to over-constrain the feasible set for the steering input. Consequently, the steering required to follow $\mathbf{ref}$ reads:
\begin{equation}
    \bar u_k = \arctan{\left(\frac{\ell_r}{\Delta s}\Delta\bar\psi_k\right)}.
\end{equation}
Thereafter, the control bounds in \eqref{eq:control_bounds} are restricted as:
\begin{equation}    \label{eq:curvature-limits}
    u_k+\bar u_k\in [u^\text{min},u^\text{max}] \;\forall k,
\end{equation}
where $u^\text{min/max} = \frac{l_r}{l_f+l_r}\delta^\text{min/max}$.

\subsubsection{Non-Linear Optimization Problem} \label{sec:optimization_problem}
Using the reformulated Frenet space bicycle kinematics model derived in Section~\ref{sec:kinematics} and the spatial corridor boundaries, $d_t^\text{lb}$ and $d_t^\text{ub}$, obtained through the BG, the path planning optimization problem is posed as follows:
\begin{align}
    \min_{u}\;\; &d^\top Q_d d + u^\top Q_u u + \lambda_\text{curve}\sum_k\tan^2 u_k\label{eq:obj} \nonumber\\ &+ \lambda_\text{risk}\sum_{k}\left(d_k - \frac{d^\text{lb}_{t_k} + d^\text{ub}_{t_k}}{2} \right)^2 \\
    \text{subject to:}\nonumber\\
    s_{k+1} &= s_k + \Delta s,\\
    d_{k+1} &= d_k + \tan(\phi_k+u_k)\Delta s,\\
    \phi_{k+1} &= \phi_k + \frac{\Delta s}{\ell_r}\frac{\sin u_k}{\cos(\phi_k+u_k)}\\
    d^\text{lb}_{t_k} &\leq d_k \leq d^\text{ub}_{t_k} \label{eq:bounds}\\
    s_0&=\hat{s}_t,~d_0=\hat{d}_t,~\phi_0=\hat{\phi}_t\\
    u_k^\text{min} &\leq u_k \leq u_k^\text{max} \label{eq:control_bounds}
\end{align}
for all $k~\in\{0,\ldots,N-1\}$ where $d=[d_1,\cdots,d_k]^\T$, $u=[u_1,\cdots,u_k]^\T$, $N$ is the number of planning steps, operator $\hat{}$ (hat) denotes the current measurement, the set $\mathbb{R}_{[u_k^\text{min},u_k^\text{max}]}$ denotes actuation limits incorporating the limits induced by $\mathbf{ref}$'s curvature \eqref{eq:curvature-limits}, and $Q$ and $\lambda$ denote penalty weight matrix and scalar, respectively. The objective function in \eqref{eq:obj} penalizes: (i) deviation from the global route (recall that $\mathbf{ref}$ corresponds to $d=0$ in the Frenet frame),  (ii) steering effort, (iii) path curvature, and (iv) distance to boundaries.

\remark
To ensure computational efficiency, the optimization problem is formulated with a convex objective function and no inequality constraints, making the kinematics model the only source of non-convexity. This is achieved by replacing the non-convex collision avoidance inequality constraints with the precomputed corridor bounds. While the kinematics model could be linearized, at the cost of model accuracy, to formulate a convex problem, we found this linearization unnecessary considering the strong computational performance demonstrated in Section \ref{sec:compuational-time-analysis}.

\subsubsection{Dynamic Obstacle Handling}   \label{sec:dynamic-obstacle-handling}

Generating boundaries for dynamic obstacles (using Algorithm~\ref{alg:boundary_generation}) may lead to recursive infeasibility as fluctuating behaviors and predictions over time may cause the upper and lower bounds to intersect. Therefore, we treat dynamic obstacles as additive risks in the optimization problem. Specifically, each predicted position over a given \emph{time} horizon is included as a convex cost in the optimization cost \eqref{eq:obj}. Therefore, the additive penalty reads:
\begin{equation}
    r_\text{dyn} = \lambda_\text{dyn} \sum_{i\in\{1,\ldots,N_t^{dyn}\}}\sum_k \;\frac{1}{(\hat{d}^{(i)}_k-d_k)^2},\label{eq:dyn_penalty}
\end{equation}
where $\lambda_\text{dyn}$ is the penalty weight, $N_t^\text{dyn}$ is the number of dynamic obstacles within the planning space at the current time $t$, and $\hat{d}^{(i)}_k$ is the predicted lateral (center of mass) position of a dynamic obstacle $i$ at step $k$.
Recall that our pipeline is based on a path-speed decomposition method (Fig.~\ref{fig:pipeline}), so the speed planner guarantees safety in space-time with respect to dynamic obstacles.

\subsubsection{Perception Noise Handling}   \label{sec:perception-noise-handling}
In the presence of perception noise, the existing optimization problem can become infeasible, especially when the ego vehicle is located close to the corridor boundaries. To ensure problem feasibility, we introduce a bounded slack variable for the bounds in \eqref{eq:bounds} as:
\begin{equation}
    d^\text{lb}_{t_k}-\alpha_{k}\leq d_k \leq d^\text{ub}_{t_k}+\alpha_{k},
\end{equation}
where $\alpha_{k}\leq \bar{\alpha}$ with a fixed $\bar{\alpha}$. Then, we add a slack penalty term in \eqref{eq:obj} as $\lambda_\alpha\sum_k \alpha_k^2$ with $\lambda_\alpha \gg 0$. 




\section{Experiments}   \label{sec:validation}
\subsection{Experimental Setup}
\begin{figure}
    \centering
    \includegraphics[width=\columnwidth]{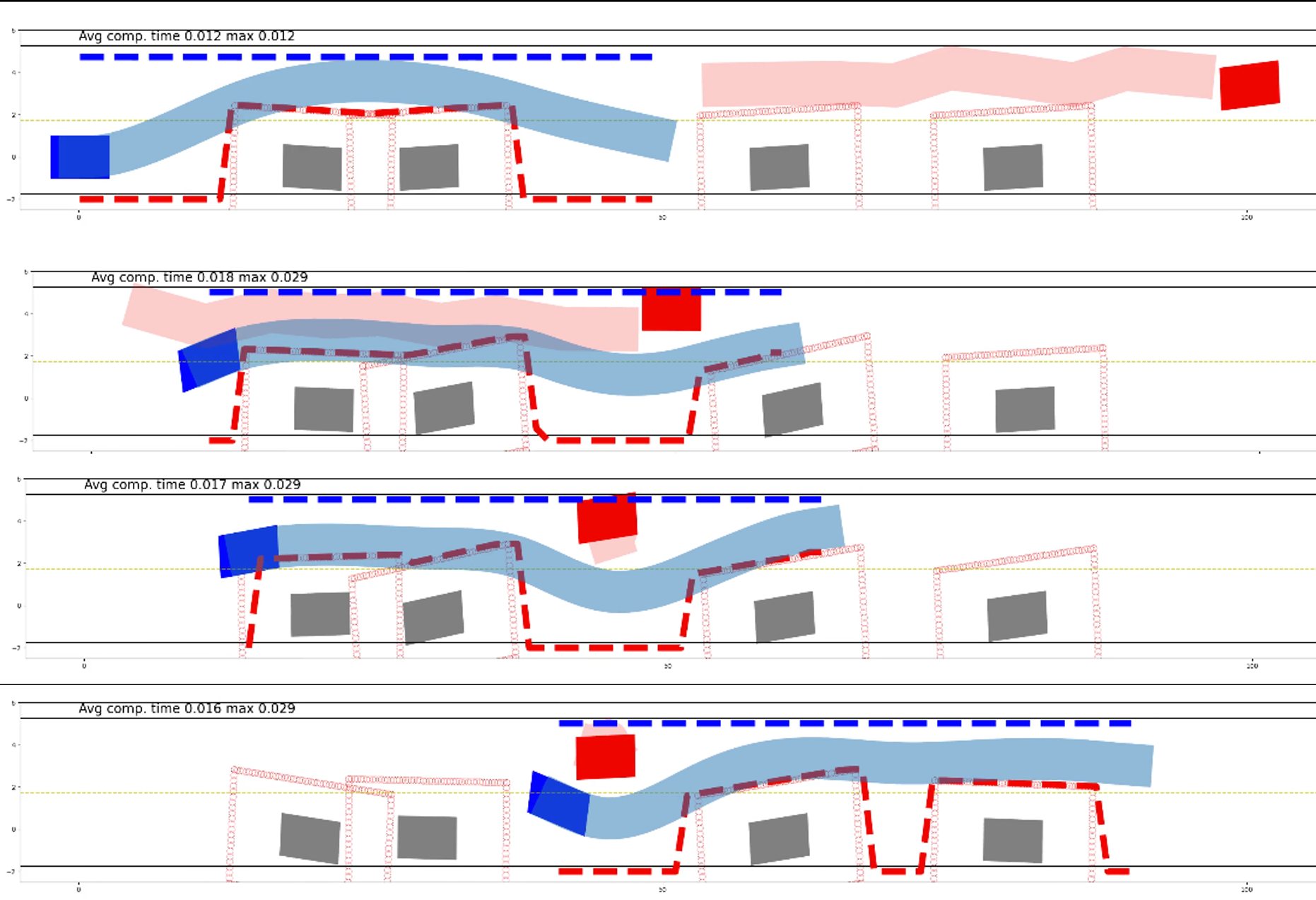}
    \caption{\textbf{Testing Scenario for Comparative Analysis}. The ego vehicle is depicted in blue, the oncoming vehicle in red, the stationary vehicles in gray, their bounding boxes with red circles, the upper/lower bound in dashed blue/red lines, the noisy perception/prediction in light red, and the ego vehicle's planned path in light blue. The scenario progression is shown from top to bottom. Note that the lower and upper boundaries are generated w.r.t. the centroid of the ego vehicle, and the ego vehicle's path is updated w.r.t. the dynamic obstacle and perception/prediction noises.}
    \label{fig:dyn}
    \vspace{-10pt}
\end{figure}

The validation studies are performed on a system running Ubuntu 22.04 LTS, equipped with an Intel® Xeon(R) Silver 4210R CPU @ 2.40GHz × 40 and an NVIDIA RTX A5000 graphics card. To assess the performance of FCP, we consider a scenario designed to replicate real-world conditions where the reference path is obstructed by stationary vehicles, requiring the ego vehicle to generate an alternative trajectory, as illustrated in Fig.~\ref{fig:dyn}. To maintain progress along its global route, the ego vehicle must navigate around the obstacles by temporarily entering the oncoming lane, return to the original lane to evade an oncoming vehicle and deviate to the oncoming lane again to finish the maneuver.

\subsection{Comparative Analysis}   \label{sec:comparative-analysis}

We consider A$^\star$ \cite{A*}, RRT$^\star$ \cite{RRT*} and Bidirectional B-RRT$^\star$ \cite{bidirectionalRRT*},\ planning algorithms as baselines to compare against FCP in the scenario depicted in Fig.~\ref{fig:dyn}.

The metrics chosen for this comparative analysis are:
\begin{enumerate*}[label=(\roman*)]
\item Algorithm runtime ($s$) - $M_t$,
\item Maximum change in yaw ($rad$) - $M_{my}$,
\item Average change in yaw ($rad$) - $M_{ay}$,
\item Average path deviation from reference path ($m$) - $M_{l}$,
\item Minimum distance to the closest vehicle ($m$) - $M_{md}$, and
\item Average distance to the closest vehicle ($m$) - $M_{ad}$.
\end{enumerate*}

These metrics are selected to assess key characteristics of the tested algorithms and the paths they generate. Specifically, $M_t$ measures computational efficiency, while $M_{l}$ quantifies the deviation of the generated path from the reference trajectory. The safety performance of the algorithm concerning the vehicles in the environment is evaluated through $M_{md}$ and $M_{ad}$. Additionally, the smoothness of the generated path is assessed using $M_{my}$ and $M_{ay}$.

The results of the quantitative comparative analysis are summarized in Table~\ref{tab:comparative-analysis}. To ensure a fair comparison, paths for the sampling-based methods, RRT$^\star$ and B-RRT$^\star$, are generated 1000 times, with metric values averaged over these runs. FCP \emph{significantly} outperforms the baseline methods in $M_t$, and $M_{my}$ highlighting its superior computational efficiency and path smoothness. FCP also outperforms the baselines in $M_{md}$ and $M_{md}$, demonstrating enhanced safety performance. In terms of $M_{ay}$, FCP surpasses the A$^\star$ and RRT$^\star$ but performs worse than B-RRT$^\star$ due to trade-offs with other critical metrics. Notably, FCP is the only algorithm that successfully passes the test scenario (without going out of bounds or colliding with any obstacles).

Qualitatively speaking, as the dynamic obstacle approaches (Fig.~\ref{fig:dyn}), the FCP path smoothly adapts to shift closer to the lower bound to account for the risk associated with the dynamic obstacle.
The path remains robust to perception noise, maintaining consistency despite fluctuating bounds. This resilience stems from IPOPT \cite{biegler2009large} navigating within the interior of the feasible region rather than along its boundaries. Additionally, the solution remains feasible even under high noise levels (second row of Fig.~\ref{fig:dyn}) due to the slack variable introduced in Section \ref{sec:perception-noise-handling}.

\begin{table}[]
\centering
\caption{Scenario-based Comparative Analysis}
\label{tab:comparative-analysis}
\begin{tabular}{|
>{\columncolor[HTML]{FFD966}}c |c|c|c|c|}
\hline
\textbf{Method/Metric} &
  \cellcolor[HTML]{FD6864}\textbf{FCP} &
  \cellcolor[HTML]{D0D0D0}\textbf{A$^\star$} &
  \cellcolor[HTML]{D0D0D0}\textbf{RRT$^\star$} &
  \cellcolor[HTML]{D0D0D0}\textbf{B-RRT$^\star$} \\ \hline
\textbf{Scenario Passed} &
  \cellcolor[HTML]{DAF2D0}Y &
  N &
  N &
  N \\ \hline
\textbf{Run-time} &
  \cellcolor[HTML]{DAF2D0}{\color[HTML]{242424} 0.035} &
  0.628 &
  0.173 &
  0.187 \\ \hline
\textbf{Max delta yaw} &
  \cellcolor[HTML]{DAF2D0}{\color[HTML]{242424} 0.053} &
  {\color[HTML]{242424} 0.785} &
  {\color[HTML]{242424} 0.816} &
  0.717 \\ \hline
\textbf{Avg delta yaw} &
  {\color[HTML]{242424} 0.016} &
  {\color[HTML]{242424} 0.051} &
  {\color[HTML]{242424} 0.003} &
  \cellcolor[HTML]{DAF2D0}0.002 \\ \hline
\textbf{Avg path div.} &
  \cellcolor[HTML]{DAF2D0}{\color[HTML]{242424} 2.336} &
  {\color[HTML]{242424} 2.360} &
  {\color[HTML]{242424} 2.867} &
  2.922 \\ \hline
\textbf{Min dist obs} &
  \cellcolor[HTML]{DAF2D0}{\color[HTML]{242424} 3.182} &
  {\color[HTML]{242424} 2.683} &
  {\color[HTML]{242424} 2.693} &
  {\color[HTML]{242424} 2.586} \\ \hline
\textbf{Avg dis obs} &
  \cellcolor[HTML]{DAF2D0}{\color[HTML]{242424} 34.99} &
  {\color[HTML]{242424} 29.90} &
  {\color[HTML]{242424} 29.81} &
  {\color[HTML]{242424} 30.22} \\ \hline
\end{tabular}
\end{table}



\begin{figure}
    \centering
    \includegraphics[width=0.7\columnwidth]{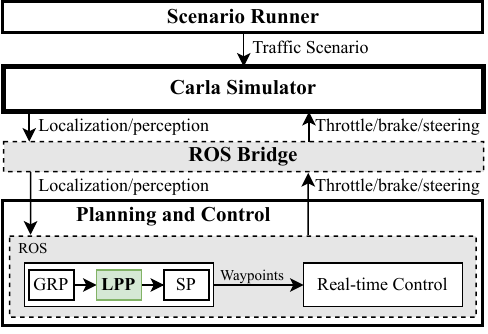}
    \caption{\textbf{CARLA Simulation Setup.} The simulation scenario, generated by the Scenario Runner, is passed on to the CARLA Simulator, which communicates with the Planning and Control ROS nodes through the ROS bridge node at a frequency of 10 Hz. GRP denotes global route planning, LPP denotes local path planning, and SP denotes speed planning.}
    \label{fig:setup}
    \vspace{-10pt}
\end{figure}

\subsection{CARLA Simulations}  \label{sec:monte-carlo}
\begin{figure}
    \centering
    \includegraphics[width=1\columnwidth]{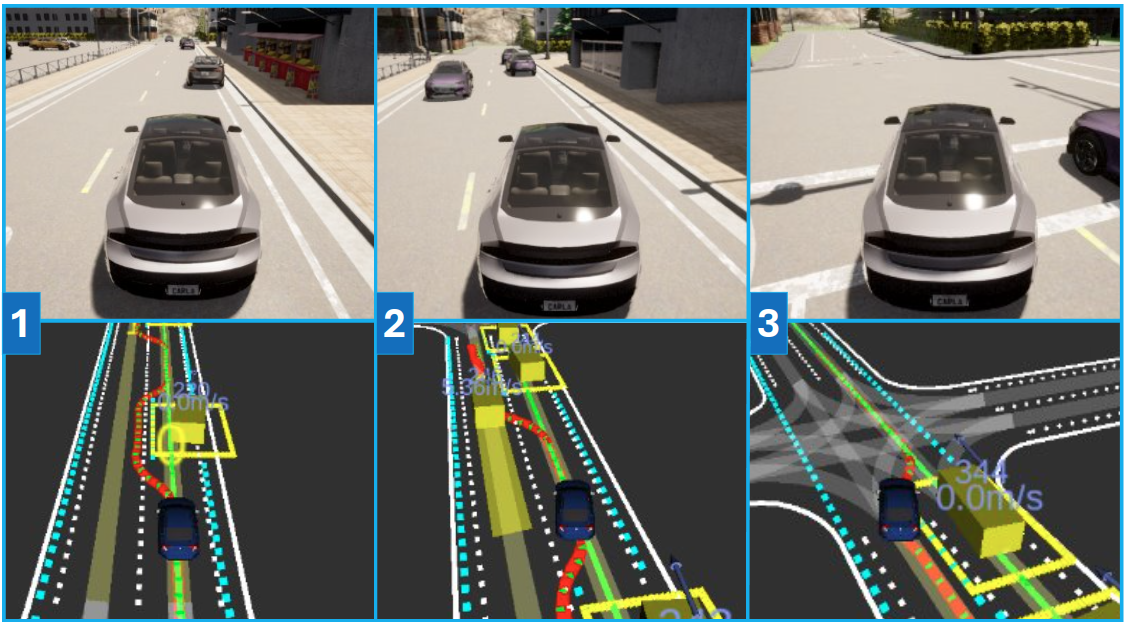}
    \caption{\textbf{CARLA Simulations.} The numbered frames show the progression of the ego vehicle through the scenario. The Rviz windows, below the Carla Pygame windows, show various objects considered during planning and the output of FCP. The ego vehicle is depicted in blue, and the obstacles are depicted as yellow cuboids surrounded by yellow safety-augmented bounding boxes. The transparent path in front of the obstacle is a constant velocity prediction path. The reference path is shown as the green line, and the local path generated by FCP is shown in red.}
    \label{fig:CARLA}
\end{figure}

To thoroughly evaluate the performance of FCP in a high-fidelity environment, the scenario shown in Fig.~\ref{fig:CARLA} is implemented in the CARLA Simulator \cite{carla}, which provides a realistic urban driving environment with sensor simulation and dynamic actors. The simulation setup within CARLA is illustrated in Fig.~\ref{fig:CARLA}. The local path planner (LPP as mentioned in Fig.~\ref{fig:setup}) employed is FCP. The other system module i.e., the speed planner (SP) is derived from \cite{anon2024mpqp}. The performance of FCP in this scenario within the CARLA environment is demonstrated in Fig.~\ref{fig:CARLA}, where the ego vehicle successfully navigates around dynamic and static obstacles.


To rigorously evaluate FCP against a standard graph-based planner, namely A$^\star$, we conduct Monte Carlo simulations for the scenario depicted in Fig.~\ref{fig:CARLA}. In these simulations, the positions and orientations of the other three vehicles are randomized within predefined ranges: the longitudinal and lateral positions vary within $10m$ and $2m$, respectively, while the vehicle headings are randomized within a $10^\circ$ range. Maintaining all system modules identical, except for the local path planning component from Fig.~\ref{fig:setup}, we run 50 simulation trials each and present the results across various performance metrics in Table~\ref{tab:monte-carlo}. Notably, none of the trials for either method resulted in a collision.

The metrics used for this Monte Carlo analysis include: \begin{enumerate*}[label=(\roman*)] \item Completion time ($s$); \item Minimum linear acceleration ($m/s^2$); \item Maximum linear acceleration ($m/s^2$); \item Minimum linear (deceleration) jerk ($m/s^3$); \item Maximum linear (acceleration) jerk ($m/s^3$); \item Maximum angular acceleration ($rad/s^2$); and, \item Maximum angular jerk ($rad/s^3$). \end{enumerate*} The linear metrics reflect the vehicle's throttle and braking behavior, primarily influenced by the speed planner, while the angular metrics correspond to steering dynamics, which are governed by the smoothness of the generated path.

The results in Table~\ref{tab:monte-carlo} demonstrate that FCP outperforms A$^\star$ in terms of path smoothness, which directly impacts the angular and steering-related metrics. A smoother path enhances passenger comfort and allows the ego vehicle to efficiently return to the reference trajectory, as indicated by lower completion time. Additionally, the smoother trajectories generated by FCP provide the speed planner with greater confidence to accelerate and decelerate along the deviation path, leading to increased linear acceleration values compared to A$^\star$. 


Furthermore, the standard deviation values reinforce the consistency of FCP’s performance relative to A$^\star$. Lower standard deviation values indicate that FCP consistently generates smooth paths across a wide range of scenarios, ensuring reliable and predictable behavior.

\begin{table}[]
\centering
\caption{Monte Carlo Simulations}
\label{tab:monte-carlo}
\resizebox{\columnwidth}{!}{%
\begin{tabular}{|clllllll|}
\hline
\rowcolor[HTML]{FFD966} 
\multicolumn{1}{|c|}{\cellcolor[HTML]{FFD966}\textbf{Model}} &
  \multicolumn{1}{c|}{\cellcolor[HTML]{FFD966}\textbf{Time}} &
  \multicolumn{1}{c|}{\cellcolor[HTML]{FFD966}\textbf{\begin{tabular}[c]{@{}c@{}}Acc.\\ Min\end{tabular}}} &
  \multicolumn{1}{c|}{\cellcolor[HTML]{FFD966}\textbf{\begin{tabular}[c]{@{}c@{}}Acc.\\ Max\end{tabular}}} &
  \multicolumn{1}{c|}{\cellcolor[HTML]{FFD966}\textbf{\begin{tabular}[c]{@{}c@{}}Jerk\\ Min\end{tabular}}} &
  \multicolumn{1}{c|}{\cellcolor[HTML]{FFD966}\textbf{\begin{tabular}[c]{@{}c@{}}Jerk\\ Max\end{tabular}}} &
  \multicolumn{1}{c|}{\cellcolor[HTML]{FFD966}\textbf{\begin{tabular}[c]{@{}c@{}}Ang.\\ Acc.\\ Max\end{tabular}}} &
  \multicolumn{1}{c|}{\cellcolor[HTML]{FFD966}\textbf{\begin{tabular}[c]{@{}c@{}}Ang.\\ Jerk\\ Max\end{tabular}}} \\ \hline
\rowcolor[HTML]{EDEDED} 
\multicolumn{8}{|c|}{\cellcolor[HTML]{EDEDED}\textbf{Average}} \\ \hline
\multicolumn{1}{|c|}{\cellcolor[HTML]{FFCCC9}\textbf{FCP}} &
  \multicolumn{1}{l|}{\cellcolor[HTML]{C6E0B4}24.743} &
  \multicolumn{1}{l|}{\cellcolor[HTML]{C6E0B4}-2.83} &
  \multicolumn{1}{l|}{1.85} &
  \multicolumn{1}{l|}{\cellcolor[HTML]{C6E0B4}-2.82} &
  \multicolumn{1}{l|}{\cellcolor[HTML]{C6E0B4}1.9} &
  \multicolumn{1}{l|}{\cellcolor[HTML]{C6E0B4}40.02} &
  \cellcolor[HTML]{C6E0B4}67.38 \\ \hline
\multicolumn{1}{|c|}{A$^\star$} &
  \multicolumn{1}{l|}{28.834} &
  \multicolumn{1}{l|}{-2.85} &
  \multicolumn{1}{l|}{\cellcolor[HTML]{C6E0B4}1.78} &
  \multicolumn{1}{l|}{-2.93} &
  \multicolumn{1}{l|}{2.48} &
  \multicolumn{1}{l|}{42.31} &
  91.1 \\ \hline
\rowcolor[HTML]{EDEDED} 
\multicolumn{8}{|c|}{\cellcolor[HTML]{EDEDED}\textbf{Standard Deviation}} \\ \hline
\multicolumn{1}{|c|}{\cellcolor[HTML]{FFCCC9}\textbf{FCP}} &
  \multicolumn{1}{l|}{\cellcolor[HTML]{C6E0B4}0.843} &
  \multicolumn{1}{l|}{\cellcolor[HTML]{C6E0B4}0.142} &
  \multicolumn{1}{l|}{0.070} &
  \multicolumn{1}{l|}{\cellcolor[HTML]{C6E0B4}0.120} &
  \multicolumn{1}{l|}{\cellcolor[HTML]{C6E0B4}0.130} &
  \multicolumn{1}{l|}{\cellcolor[HTML]{C6E0B4}1.855} &
  \cellcolor[HTML]{C6E0B4}4.235 \\ \hline
\multicolumn{1}{|c|}{A$^\star$} &
  \multicolumn{1}{l|}{2.081} &
  \multicolumn{1}{l|}{0.221} &
  \multicolumn{1}{l|}{\cellcolor[HTML]{C6E0B4}0.066} &
  \multicolumn{1}{l|}{0.124} &
  \multicolumn{1}{l|}{0.200} &
  \multicolumn{1}{l|}{3.348} &
  11.685 \\ \hline
\end{tabular}%
}
\vspace{-10pt}
\end{table}

\subsection{Computational Time Analysis}    \label{sec:compuational-time-analysis}
\begin{figure}
    \centering
    \includegraphics[width=0.7\columnwidth]{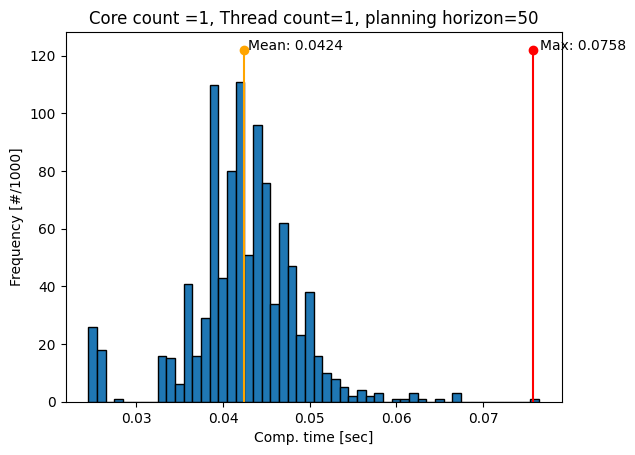}
    \vspace{-3mm}
    \caption{\textbf{Computational Efficiency Analysis}. The test scenario in Fig.~\ref{fig:dyn} is randomized, similar to the Monte-Carlo simulations in Section~\ref{sec:monte-carlo}, and repeated 1000 times to obtain the FCP computation time distribution.}
    \label{fig:comp-time}
\end{figure}

We assess the computational efficiency of our algorithm, implemented using IPOPT in Casadi (Python), by analyzing the algorithm runtime distribution across 1000 randomized test scenario runs, with both core and thread counts restricted to one. The runtime distribution is illustrated in Fig.~\ref{fig:comp-time}. Our algorithm achieves an average computation time of 0.0424 seconds and a maximum of 0.0758 seconds, demonstrating its capability for real-time operation. Furthermore, even greater computational efficiency could be achieved if implemented in a compiled language such as C/C++.

\subsection{Hardware Demonstration}

\begin{figure}
    \centering
    \includegraphics[width=0.95\columnwidth]{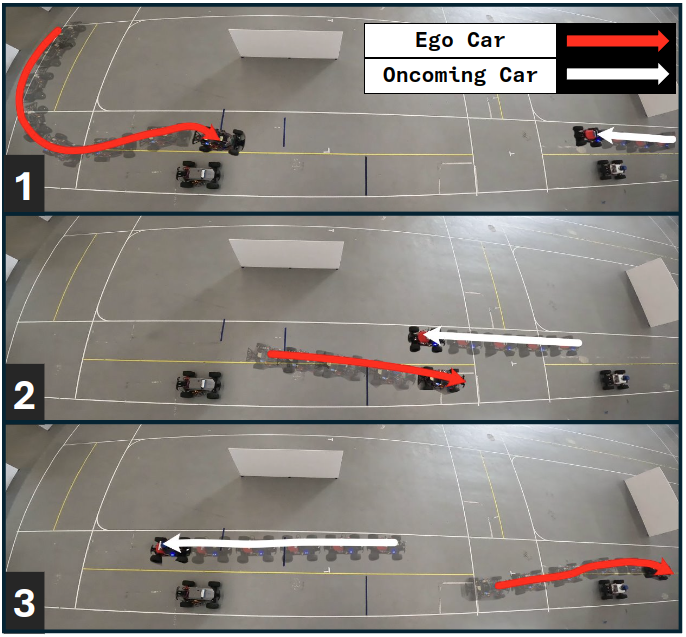}
    \caption{\textbf{Hardware Demonstration.} A scenario involving two parked vehicles and an oncoming vehicle is demonstrated using 1/10-scale autonomous cars. The numbered frames illustrate the progression of the scenario, with each frame containing snapshots of the robots on the test track along with super-imposed trajectories depicting the paths taken by the vehicles.}
    \label{fig:hardware-demo}
    \vspace{-15pt}
\end{figure}

We demonstrate the performance of FCP in a scenario involving four customized 1/10-scale Multi-agent System for non-Holonomic Racing (MuSHR) \cite{mushr} autonomous vehicles—comprising one ego vehicle, two stationary vehicles mimicking parked vehicles, and one moving vehicle—at our testing facility located at Honda Research Institute Inc USA, San Jose, CA. This scenario is similar to the simulation scenario shown in Fig.~\ref{fig:CARLA}, with an added layer of difficulty for FCP to encounter a static obstacle on a curved path as soon as it turns left. The Robot Operating System (ROS) serves as the communication framework, facilitating interaction between sensors, actuators, and computing units, and is executed within a Docker container. FCP runs inside a separate Docker container, ensuring efficient execution. The ego vehicle utilizes LiDAR based localization to estimate its state, including position, velocity, and orientation, using a predefined grid map of the track and surrounding landmarks. The planner operates at a frequency of 10Hz and runs on an Intel NUC mini PC onboard the MuSHR vehicle, which has Ubuntu 20.04 LTS as its operating system (OS). The ego vehicle’s trajectory is visualized through sequentially numbered frames in Fig.~\ref{fig:hardware-demo} with the paths taken by the ego and the dynamic obstacle superimposed in each frame. The successful deployment of FCP on the physical robot, along with its smooth execution, highlights its computational efficiency and robustness in handling real-world uncertainties, perception noises and delays.



\section{Conclusion}
We propose a computationally efficient risk-aware local path planning algorithm to generate smooth deviation paths in reference to a fixed path for automated driving applications.
Validation results from CARLA simulations, comparative analysis and scaled autonomous vehicle tests demonstrate the effectiveness of our approach.











\bibliographystyle{IEEEtran}
\bibliography{ref}

\begin{thebibliography}{10}
\providecommand{\url}[1]{#1}
\csname url@samestyle\endcsname
\providecommand{\newblock}{\relax}
\providecommand{\bibinfo}[2]{#2}
\providecommand{\BIBentrySTDinterwordspacing}{\spaceskip=0pt\relax}
\providecommand{\BIBentryALTinterwordstretchfactor}{4}
\providecommand{\BIBentryALTinterwordspacing}{\spaceskip=\fontdimen2\font plus
\BIBentryALTinterwordstretchfactor\fontdimen3\font minus \fontdimen4\font\relax}
\providecommand{\BIBforeignlanguage}[2]{{%
\expandafter\ifx\csname l@#1\endcsname\relax
\typeout{** WARNING: IEEEtran.bst: No hyphenation pattern has been}%
\typeout{** loaded for the language `#1'. Using the pattern for}%
\typeout{** the default language instead.}%
\else
\language=\csname l@#1\endcsname
\fi
#2}}
\providecommand{\BIBdecl}{\relax}
\BIBdecl

\bibitem{kant1986toward}
K.~Kant and S.~W. Zucker, ``Toward efficient trajectory planning: The path-velocity decomposition,'' \emph{The international journal of robotics research}, vol.~5, no.~3, pp. 72--89, 1986.

\bibitem{fraichard1993dynamic}
T.~Fraichard and C.~Laugier, ``Dynamic trajectory planning, path-velocity decomposition and adjacent paths,'' in \emph{IJCAI}, 1993, pp. 1592--1599.

\bibitem{anon2024mpqp}
A.~M. A{\~n}on, S.~Bae, M.~Saroya, and D.~Isele, ``Multi-profile quadratic programming (mpqp) for optimal gap selection and speed planning of autonomous driving,'' \emph{arXiv preprint arXiv:2401.06305}, 2024.

\bibitem{koenig2002d}
S.~Koenig and M.~Likhachev, ``D* lite,'' in \emph{Eighteenth national conference on Artificial intelligence}, 2002, pp. 476--483.

\bibitem{likhachev2005anytime}
M.~Likhachev, D.~I. Ferguson, G.~J. Gordon, A.~Stentz, and S.~Thrun, ``Anytime dynamic a*: An anytime, replanning algorithm.'' in \emph{ICAPS}, vol.~5, 2005, pp. 262--271.

\bibitem{multifuture}
D.~Isele, A.~M. Anon, F.~M. Tariq, G.~Yeh, A.~Singh, and S.~Bae, ``Delayed-decision motion planning in the presence of multiple predictions,'' \emph{arXiv preprint arXiv:2502.20636}, 2025.

\bibitem{huang2023neural}
Z.~Huang, H.~Chen, and K.~Driggs-Campbell, ``Neural informed rrt* with point-based network guidance for optimal sampling-based path planning,'' \emph{arXiv preprint arXiv:2309.14595}, 2023.

\bibitem{anjian_consistencyModel}
A.~Li, S.~Bae, D.~Isele, R.~Beeson, and F.~M. Tariq, ``End-to-end predictive planner for autonomous driving with consistency models,'' \emph{arXiv preprint arXiv:2502.08033}, 2025.

\bibitem{hwang1992potential}
Y.~K. Hwang, N.~Ahuja \emph{et~al.}, ``A potential field approach to path planning.'' \emph{IEEE transactions on robotics and automation}, vol.~8, no.~1, pp. 23--32, 1992.

\bibitem{A*}
P.~E. Hart, N.~J. Nilsson, and B.~Raphael, ``A formal basis for the heuristic determination of minimum cost paths,'' \emph{IEEE transactions on Systems Science and Cybernetics}, vol.~4, no.~2, pp. 100--107, 1968.

\bibitem{daniel2010theta}
K.~Daniel, A.~Nash, S.~Koenig, and A.~Felner, ``Theta*: Any-angle path planning on grids,'' \emph{Journal of Artificial Intelligence Research}, vol.~39, pp. 533--579, 2010.

\bibitem{RRT*}
S.~Karaman and E.~Frazzoli, ``Incremental sampling-based algorithms for optimal motion planning,'' 2011.

\bibitem{jordan2013optimal}
M.~Jordan and A.~Perez, ``Optimal bidirectional rapidly-exploring random trees,'' 2013.

\bibitem{xu2012real}
W.~Xu, J.~Wei, J.~M. Dolan, H.~Zhao, and H.~Zha, ``A real-time motion planner with trajectory optimization for autonomous vehicles,'' in \emph{2012 IEEE International Conference on Robotics and Automation}.\hskip 1em plus 0.5em minus 0.4em\relax IEEE, 2012, pp. 2061--2067.

\bibitem{dirckx2023optimal}
D.~Dirckx, M.~Bos, W.~Decr{\'e}, and J.~Swevers, ``Optimal and reactive control for agile drone flight in cluttered environments,'' \emph{IFAC-PapersOnLine}, vol.~56, no.~2, pp. 6273--6278, 2023.

\bibitem{kong2015kinematic}
J.~Kong, M.~Pfeiffer, G.~Schildbach, and F.~Borrelli, ``Kinematic and dynamic vehicle models for autonomous driving control design,'' in \emph{2015 IEEE intelligent vehicles symposium (IV)}.\hskip 1em plus 0.5em minus 0.4em\relax IEEE, 2015, pp. 1094--1099.

\bibitem{rcms}
F.~M. Tariq, D.~Isele, J.~S. Baras, and S.~Bae, ``Rcms: Risk-aware crash mitigation system for autonomous vehicles,'' in \emph{2023 IEEE 26th International Conference on Intelligent Transportation Systems (ITSC)}.\hskip 1em plus 0.5em minus 0.4em\relax IEEE, 2023, pp. 3950--3957.

\bibitem{phan2020covernet}
T.~Phan-Minh, E.~C. Grigore, F.~A. Boulton, O.~Beijbom, and E.~M. Wolff, ``Covernet: Multimodal behavior prediction using trajectory sets,'' in \emph{Proceedings of the IEEE/CVF Conference on Computer Vision and Pattern Recognition}, 2020, pp. 14\,074--14\,083.

\bibitem{salzmann2020trajectron++}
T.~Salzmann, B.~Ivanovic, P.~Chakravarty, and M.~Pavone, ``Trajectron++: Dynamically-feasible trajectory forecasting with heterogeneous data,'' in \emph{Computer Vision--ECCV 2020: 16th European Conference, Glasgow, UK, August 23--28, 2020, Proceedings, Part XVIII 16}.\hskip 1em plus 0.5em minus 0.4em\relax Springer, 2020, pp. 683--700.

\bibitem{sadigh2016planning}
D.~Sadigh, S.~Sastry, S.~A. Seshia, and A.~D. Dragan, ``Planning for autonomous cars that leverage effects on human actions.'' in \emph{Robotics: Science and systems}, vol.~2.\hskip 1em plus 0.5em minus 0.4em\relax Ann Arbor, MI, USA, 2016, pp. 1--9.

\bibitem{ma2021reinforcement}
X.~Ma, J.~Li, M.~J. Kochenderfer, D.~Isele, and K.~Fujimura, ``Reinforcement learning for autonomous driving with latent state inference and spatial-temporal relationships,'' in \emph{2021 IEEE International Conference on Robotics and Automation (ICRA)}.\hskip 1em plus 0.5em minus 0.4em\relax IEEE, 2021, pp. 6064--6071.

\bibitem{bae2022lane}
S.~Bae, D.~Isele, A.~Nakhaei, P.~Xu, A.~M. A{\~n}on, C.~Choi, K.~Fujimura, and S.~Moura, ``Lane-change in dense traffic with model predictive control and neural networks,'' \emph{IEEE Transactions on Control Systems Technology}, vol.~31, no.~2, pp. 646--659, 2022.

\bibitem{hu2023active}
H.~Hu, D.~Isele, S.~Bae, and J.~F. Fisac, ``Active uncertainty reduction for safe and efficient interaction planning: A shielding-aware dual control approach,'' \emph{The International Journal of Robotics Research}, p. 02783649231215371, 2023.

\bibitem{bidirectional_overtaking}
F.~M. Tariq, N.~Suriyarachchi, C.~Mavridis, and J.~S. Baras, ``Autonomous vehicle overtaking in a bidirectional mixed-traffic setting,'' in \emph{2022 American Control Conference (ACC)}.\hskip 1em plus 0.5em minus 0.4em\relax IEEE, 2022, pp. 3132--3139.

\bibitem{frenet1852courbes}
F.~Frenet, ``Sur les courbes {\`a} double courbure,'' \emph{Journal de math{\'e}matiques pures et appliqu{\'e}es}, vol.~17, pp. 437--447, 1852.

\bibitem{tariq2022slas}
F.~M. Tariq, D.~Isele, J.~S. Baras, and S.~Bae, ``Slas: Speed and lane advisory system for highway navigation,'' in \emph{2022 IEEE 61st Conference on Decision and Control (CDC)}.\hskip 1em plus 0.5em minus 0.4em\relax IEEE, 2022, pp. 6979--6986.

\bibitem{carla}
A.~Dosovitskiy, G.~Ros, F.~Codevilla, A.~Lopez, and V.~Koltun, ``Carla: An open urban driving simulator,'' in \emph{Conference on robot learning}.\hskip 1em plus 0.5em minus 0.4em\relax PMLR, 2017, pp. 1--16.

\bibitem{vehicleRotationModel}
T.~Tashiro, ``Vehicle steering control with mpc for target trajectory tracking of autonomous reverse parking,'' in \emph{2013 ieee international conference on control applications (cca)}.\hskip 1em plus 0.5em minus 0.4em\relax IEEE, 2013, pp. 247--251.

\bibitem{dbscan}
M.~Ester, H.-P. Kriegel, J.~Sander, X.~Xu \emph{et~al.}, ``A density-based algorithm for discovering clusters in large spatial databases with noise,'' in \emph{kdd}, vol.~96, no.~34, 1996, pp. 226--231.

\bibitem{song2015decision}
Y.-Y. Song and L.~Ying, ``Decision tree methods: applications for classification and prediction,'' \emph{Shanghai archives of psychiatry}, vol.~27, no.~2, p. 130, 2015.

\bibitem{planningSurvey}
O.~Sharma, N.~C. Sahoo, and N.~B. Puhan, ``Recent advances in motion and behavior planning techniques for software architecture of autonomous vehicles: A state-of-the-art survey,'' \emph{Engineering applications of artificial intelligence}, vol. 101, p. 104211, 2021.

\bibitem{bicycleModel}
J.~Kong, M.~Pfeiffer, G.~Schildbach, and F.~Borrelli, ``Kinematic and dynamic vehicle models for autonomous driving control design,'' in \emph{2015 IEEE intelligent vehicles symposium (IV)}.\hskip 1em plus 0.5em minus 0.4em\relax IEEE, 2015, pp. 1094--1099.

\bibitem{bidirectionalRRT*}
M.~Jordan and A.~Perez, ``Optimal bidirectional rapidly-exploring random trees,'' 2013.

\bibitem{biegler2009large}
L.~T. Biegler and V.~M. Zavala, ``Large-scale nonlinear programming using ipopt: An integrating framework for enterprise-wide dynamic optimization,'' \emph{Computers \& Chemical Engineering}, vol.~33, no.~3, pp. 575--582, 2009.

\bibitem{mushr}
S.~S. Srinivasa \emph{et~al.}, ``Mushr: A low-cost, open-source robotic racecar for education and research,'' \emph{arXiv preprint arXiv:1908.08031}, 2019.

\end{thebibliography}

\end{document}